\documentclass[10pt,twocolumn,letterpaper]{article}

\usepackage[pagenumbers]{cvpr}
\usepackage{colortbl}
\usepackage{placeins}
\usepackage{stfloats}
\usepackage{fontawesome5}

\definecolor{cvprblue}{rgb}{0.21,0.49,0.74}
\usepackage[breaklinks,colorlinks,allcolors=cvprblue]{hyperref}
\hypersetup{
    pdftitle={Reading, Not Adapting: Unlocking Frozen Visual Knowledge for AI-Generated Image Detection},
    pdfauthor={Yuncheng Guo, Ying Dong, Xiaodong Gu}
}

\title{Reading, Not Adapting: Unlocking Frozen Visual Knowledge\\
for AI-Generated Image Detection}

\author{Yuncheng Guo\textsuperscript{1}, Ying Dong,
Xiaodong Gu\textsuperscript{2}\thanks{Corresponding author}\\
Department of Electronic Engineering, Fudan University\\
{\small\href{https://github.com/yunncheng/ReMoT}{\faGithub\enspace\nolinkurl{https://github.com/yunncheng/ReMoT}}}\\
{\small\textsuperscript{1}\href{mailto:guoyc19@fudan.edu.cn}{\texttt{guoyc19@fudan.edu.cn}},\quad
\textsuperscript{2}\href{mailto:xdgu@fudan.edu.cn}{\texttt{xdgu@fudan.edu.cn}}}}

\begin{document}
\maketitle
\begin{abstract}
Generalization to unseen image generators has driven efforts
to adapt pretrained visual representations for detection.
Yet recent findings show that vision foundation models already
encode transferable detection knowledge, while further
adaptation can impair generalization.
This motivates a shift from backbone adaptation
to knowledge readout.
Conventional linear probing offers a direct way to read this knowledge
but is limited to final-layer RGB features. Multi-cue detectors broaden
the evidence base, but their designs introduce other tradeoffs: adaptation may compromise pretrained transferability,
heterogeneous encoders complicate feature integration, and late fusion
precludes intermediate cross-view interaction.
To overcome these limitations, we propose \textbf{ReMoT},
a Mixture-of-Transformers-inspired framework that reads
complementary knowledge from spatially aligned RGB and wavelet
views within a fully frozen visual foundation. ReMoT employs staged
joint attention to enable controlled local and global
cross-view interaction, while learnable fusion queries
progressively read complementary evidence across
Transformer depth without writeback.
ReMoT achieves \textbf{91.2\% average accuracy} across eight
benchmark groups with 3,074 trainable parameters,
attaining the highest overall average among
the evaluated methods. These results support structured
knowledge readout as a promising alternative to backbone adaptation.
\end{abstract}

\section{Introduction}
\label{sec:intro}

The rapid evolution of image generation poses a fundamental
challenge to AI-generated image detection: generalizing
beyond the generators encountered during training.
Detectors must recognize unseen synthesis methods~\cite{ref30} while
remaining robust to variations in image content, source
distributions, and post-processing~\cite{li2025realworld}.
Yet learned forensic cues~\cite{cnnspot2020,ref10,ref8}
may capture generator-specific artifacts or dataset
biases rather than transferable evidence of synthesis
~\cite{zhu2023genimage,chen2025dda,park2025community,
ref12}. UnivFD~\cite{ojha2023universal}
provided early evidence of the potential of pretrained
visual knowledge, demonstrating transferable detection
using frozen CLIP~\cite{radford2021clip} representations
and simple classifiers. Motivated by this potential,
considerable research has focused on adapting pretrained
backbones to improve cross-generator generalization.
Methods such as FatFormer~\cite{ref0} and Effort~\cite{yan2025effort}
modify pretrained feature processing for detection while seeking
to preserve transferable visual knowledge.

\begin{figure}[t]
    \centering
    \includegraphics[width=\linewidth]{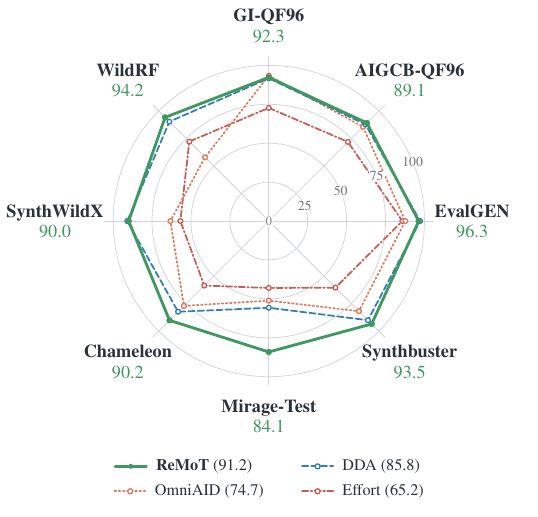}
    \caption{\textbf{Generalization across eight benchmarks.} Axes report balanced accuracy;
    parentheses give eight-group averages.}
    \label{fig:radar}
\end{figure}

\begin{figure*}[t]
    \centering
    \includegraphics[width=\textwidth]{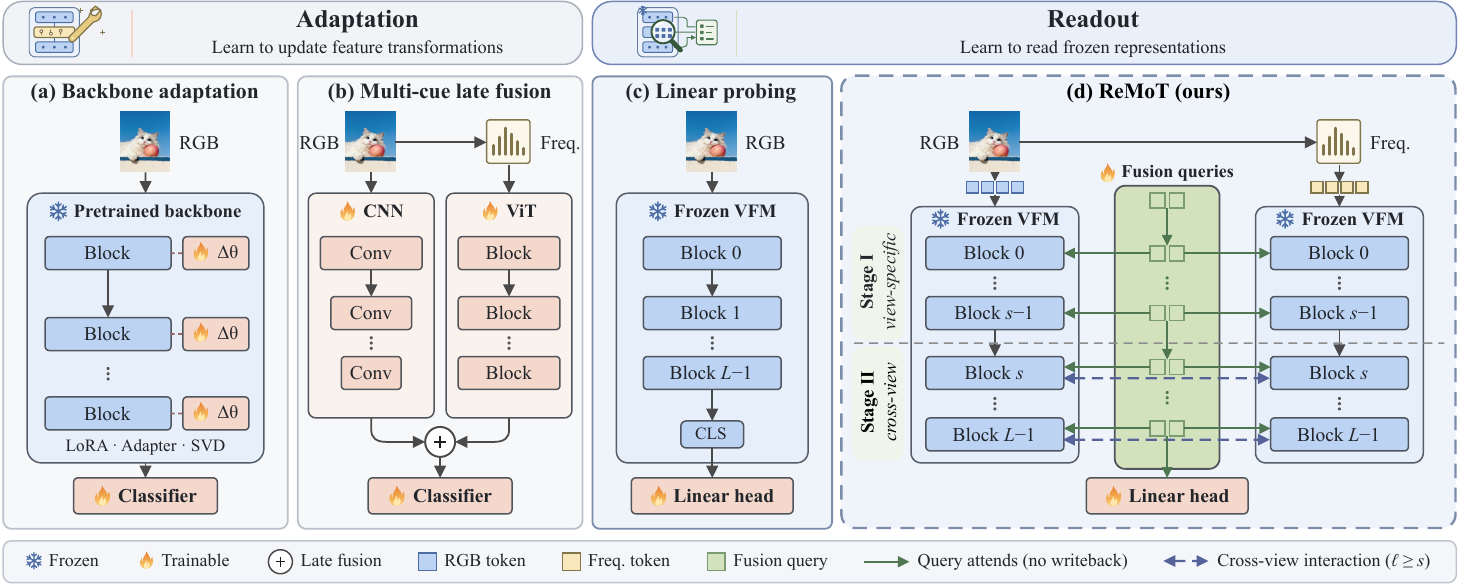}
    \caption{\textbf{From backbone adaptation to knowledge readout.}
    Representative designs: (A) task-trained backbone adaptation;
    (B) heterogeneous encoding with late fusion; (C) final-layer RGB
    linear probing; (D) frozen dual-view joint processing with staged
    interaction and progressive query readout.}
    \label{fig:teaser}
\end{figure*}

Recent findings, however, challenge whether such
adaptation is always necessary. \emph{Simplicity Prevails}
~\cite{zhou2026simplicity} demonstrates that modern
vision foundation models (VFMs), particularly DINOv3
~\cite{simeoni2025dinov3}, already encode highly
transferable detection knowledge accessible through
simple linear probing. Moreover, additional adaptation,
including LoRA~\cite{hu2022lora}, can substantially
degrade generalization in its evaluated settings.
These findings motivate a shift in emphasis from
\emph{backbone adaptation} to \emph{knowledge readout}.
Yet how to effectively exploit this knowledge remains
underexplored. Conventional frozen linear probing
typically relies on a single final-layer RGB
representation, leaving complementary evidence across
views and Transformer depth largely unexploited.
Meanwhile, multi-cue detectors incorporate semantic,
spatial, and frequency-related evidence, but representative
approaches often involve adaptation, heterogeneous
encoding, and late fusion
~\cite{yan2025sanity,cheng2025cospy}.
These designs entail distinct tradeoffs: adaptation
may favor training-specific patterns; heterogeneous
encoders produce representations with different
inductive biases and abstraction levels, complicating
their integration; and late fusion prevents intermediate
interaction between independently encoded views.
This leaves a fundamental question:
\emph{how can frozen encoders with the same architecture
jointly exploit complementary forensic evidence beyond
final-layer RGB probing?}

Our central insight is that \emph{organizing complementary evidence
and learning how to read it can better exploit pretrained detection
knowledge without learning task-specific visual transformations}.
Complementary views can interact
within a common pretrained architecture, while their evidence
can be accessed throughout feature extraction. Cross-view interaction
and readout determine how this knowledge is accessed.
Effective readout therefore involves both cross-view organization
and access across Transformer depth.
We realize this perspective in \textbf{ReMoT}
(\textbf{Re}adout with \textbf{M}ixture-\textbf{o}f-\textbf{T}ransformers),
a framework for progressive cross-view readout in
AI-generated image detection.
MoT~\cite{liang2025mot} combines modality-specific
Transformer parameters with joint attention across modalities.
ReMoT adopts this principle for spatially aligned RGB
and discrete wavelet transform (DWT) views. Both streams use
the same frozen DINOv3 architecture and pretrained weights, providing a common foundation
for MoT-inspired joint attention.
A \emph{staged joint-attention} mechanism initially
preserves view-specific processing and subsequently
enables controlled cross-view interaction through
local patch correspondence and global register-mediated
communication. Meanwhile, learnable fusion queries
\emph{progressively read} both views across Transformer
depth using frozen attention projections, without
writing back to view tokens. Together, these mechanisms organize
cross-view evidence for progressive readout while preserving
pretrained backbone weights. Trained solely
on GenImage Stable Diffusion v1.4 (SD~v1.4), ReMoT achieves
\textbf{91.2\% average accuracy} across eight
benchmark groups (Figure~\ref{fig:radar}) with only \textbf{3,074 trainable
parameters} in its equivalent implementation,
attaining the highest overall average among the
evaluated methods. An independently trained variant without
direct cross-view interaction reaches 87.0\%, highlighting the
value of evidence exchange alongside progressive query readout.

\begin{figure*}[t!]
    \centering
    \includegraphics[width=\textwidth]{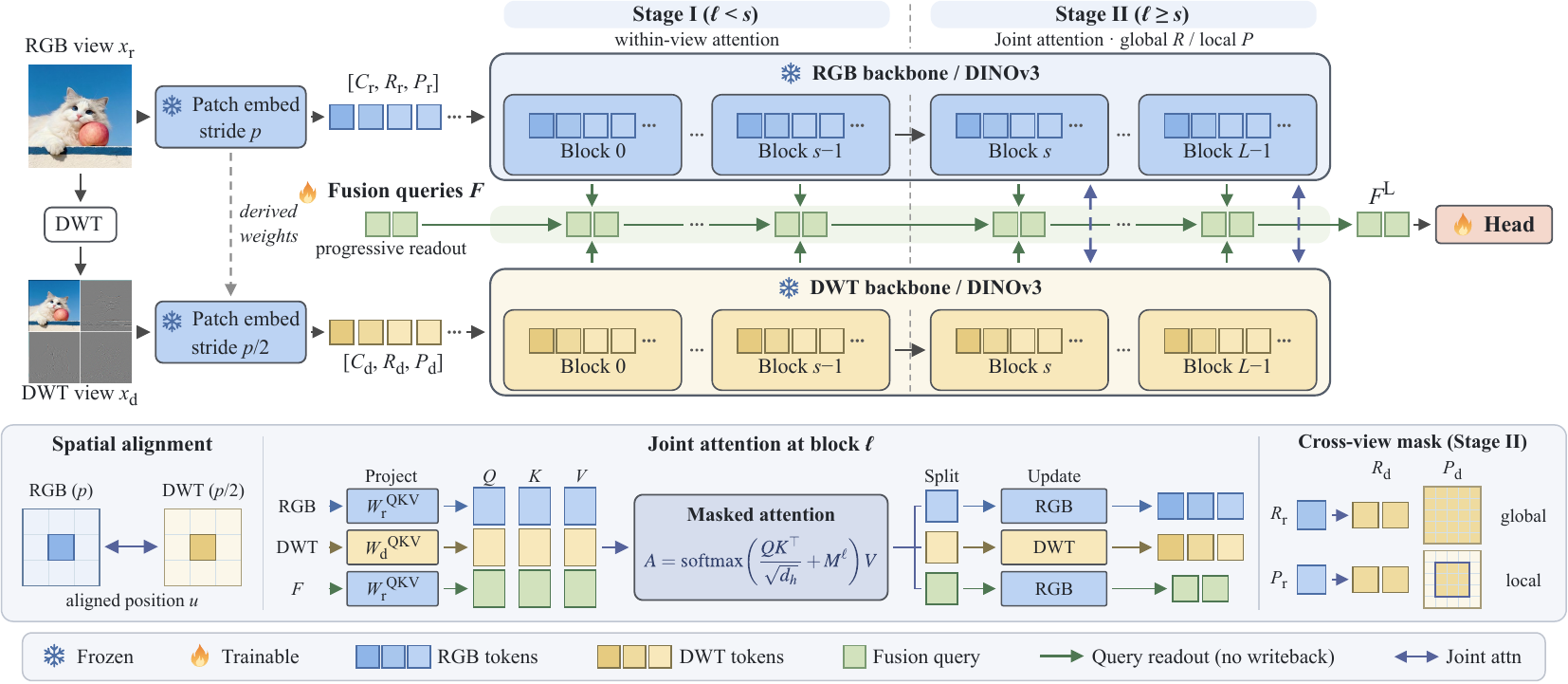}
    \caption{\textbf{ReMoT framework.}
    Frozen RGB/DWT streams interact from block $s$; $K$ fusion queries
    progressively read both without writeback. Insets show spatial
    alignment, joint attention, and Stage II connectivity
    (Figure~\ref{fig:mask}). Only fusion-query embeddings and the classification head
    are trainable. DWT tiles are individually rescaled for display.}
    \label{fig:method}
\end{figure*}

Our contributions are summarized as follows:
\begin{enumerate}[nosep, leftmargin=*]
    \item We advance a \textbf{readout-centered perspective}
    for generalizable detection, shifting
    the design focus from adapting pretrained representations
    to organizing and accessing their existing detection knowledge.

    \item We propose \textbf{ReMoT}, combining frozen RGB/DWT
    streams, staged MoT-inspired joint attention with local patch
    and global register exchange, and learnable fusion queries
    for progressive readout across depth without writing back
    to either visual stream.

    \item ReMoT achieves a leading \textbf{91.2\% average accuracy}
    across eight benchmark groups, outperforming prior
    adaptation-based detectors and showing the potential of
    \textbf{knowledge readout} for generalizable detection.
\end{enumerate}

\section{Related work}
\label{sec:related}

\subsection{From adaptation to readout}

Generator-specific artifacts and dataset biases limit
cross-generator detection~\cite{chen2025dda,ref12}.
UnivFD~\cite{ojha2023universal} demonstrated the value of
pretrained knowledge through simple classifiers on frozen
CLIP~\cite{radford2021clip} features. Self-supervised
foundations offer another source of transferable visual
features~\cite{oquab2024dinov2}. Subsequent adaptation
methods seek stronger detection representations while
preserving transferability: FatFormer~\cite{ref0} introduces
forgery-aware transformations, and Effort~\cite{yan2025effort}
learns in an orthogonal subspace while freezing principal
pretrained components. Prompt-based methods likewise adapt
pretrained representations for detection~\cite{tan2024c2p,ref38}.
OmniAID~\cite{guo2025omniaid} decouples content-dependent
semantic flaws from universal artifacts through routable
semantic experts and an always-active artifact expert.
Complementary reconstruction-based training improves cross-generator
generalization~\cite{chen2024drct,rajan2025aligned}; DDA~\cite{chen2025dda}
aligns real and synthetic images in pixel and frequency
domains to suppress spurious correlations.
Recent evidence questions whether further adaptation is
necessary. \emph{Simplicity Prevails}~\cite{zhou2026simplicity}
finds that modern VFMs, including DINOv3~\cite{simeoni2025dinov3},
already support strong generalization through linear probing,
while additional adaptation can degrade performance in its
evaluated settings. These findings motivate readout beyond
final-layer probing: ReMoT explores how existing visual knowledge
can be accessed across views and depth while backbone weights remain frozen.

\subsection{Cross-view evidence integration}

Beyond final-layer RGB probing, complementary evidence can be
sought across image views and Transformer depth.
Local artifact modeling~\cite{ref1,li2024safe,ref19,ref23},
reconstruction~\cite{ref18}, and quantization cues~\cite{ref24}
complement RGB appearance.
Spectral analyses caution against universal fingerprints
~\cite{chandrasegaran2021closer,dong2022think}, motivating
robust spectral representations~\cite{ref7,ref37,ref47}.
Layer Consistency Matters~\cite{ref39} identifies evidence
in latent transitions across layers. Multi-cue detectors and
multimodal reasoning~\cite{ref31} explore evidence integration:
AIDE~\cite{yan2025sanity} integrates semantic features with
forensic features from texture-selected patches, while
CO-SPY~\cite{cheng2025cospy} enhances and adaptively fuses
semantic and pixel-level features. However, heterogeneous
encoders complicate feature alignment, and late fusion limits
exchange during feature extraction. Developed for multimodal
pretraining, MoT~\cite{liang2025mot} combines trainable
modality-specific Transformer parameters with joint self-attention.
ReMoT borrows this joint-processing principle to exchange and
progressively read spatially aligned RGB and wavelet
evidence~\cite{mallat1989wavelet} using frozen pretrained backbones.

\section{Method}
\label{sec:method}

ReMoT organizes complementary evidence within a frozen visual
foundation through MoT-inspired joint processing, staged cross-view
interaction, and progressive query readout (Figure~\ref{fig:method}).
The two views share a joint attention space, with a stage-dependent mask
governing their interaction. Learnable queries collect evidence
across Transformer depth while pretrained backbone weights remain frozen.

\subsection{MoT-inspired frozen cross-view architecture}
\label{sec:method-architecture}

\textbf{Complementary view construction.}
Given an RGB image $x\in\mathbb{R}^{3\times H\times W}$,
we construct an appearance view $x_r=x$ and a
spatial--frequency view using a one-level,
channel-wise discrete wavelet transform (DWT)~\cite{mallat1989wavelet}:
\begin{equation}
    x_d=\operatorname{DWT}(x)
    =[x_{LL},x_{LH},x_{HL},x_{HH}].
    \label{eq:wavelet-view}
\end{equation}
Cropping the subbands to $H/2\times W/2$ and concatenating them
gives 12 spatially indexed channels of low-frequency content and
directional detail. Separate frozen patch stems embed RGB and DWT:
the former inherits DINOv3 weights~\cite{simeoni2025dinov3}, and the
latter is deterministically derived from them. Strides $p$ and $p/2$
align the original and half-resolution views' patch grids.

\textbf{MoT-inspired joint processing.}
For each view $m\in\{r,d\}$, let $C_m$, $R_m$,
and $P_m$ denote its class, register~\cite{darcet2024registers}, and patch
tokens, with $U_m=[R_m,P_m]$. Appending $K$
learnable fusion queries $F=[F_1,\ldots,F_K]$
forms
\begin{equation}
    Z^0=[C_r,U_r,C_d,U_d,F].
    \label{eq:token-sequence}
\end{equation}
Class and register embeddings remain frozen.

Following Mixture-of-Transformers (MoT)
~\cite{liang2025mot}, built on the Transformer
architecture~\cite{vaswani2017attention,dosovitskiy2021vit}, each view uses its own
Transformer parameter slot; fusion queries use the RGB slot.
The $L$ blocks are indexed by $\ell=0,\ldots,L-1$.
Each block projects token state $z$ using its assigned slot $m$:
\begin{equation}
    (q,k,v)=W_m^{QKV}\operatorname{LN}_m(z),
    \label{eq:mot-projection}
\end{equation}
where the block index, biases, and head indices are omitted for
clarity; each block retains its own pretrained parameters.
Patch queries and keys receive pretrained 2D RoPE with aligned
coordinates across views~\cite{su2024roformer}; non-patch tokens
are not rotated. For each attention head of width $d_h$,
concatenate all token projections along the token axis to obtain
$\mathbf{Q},\mathbf{K},\mathbf{V}$ and compute
\begin{equation}
    A=\operatorname{softmax}\left(
        \frac{\mathbf{Q}\mathbf{K}^\top}{\sqrt{d_h}}+M^\ell
    \right)\mathbf{V}.
    \label{eq:joint-attention}
\end{equation}
Output projections, normalization, and feed-forward operations
follow the corresponding slots and pretrained residual structure.
Both parameter slots remain frozen at their shared DINOv3 initialization.

\subsection{Staged joint attention}
\label{sec:method-staged}

Unrestricted cross-view attention may mix
complementary evidence before view-specific
processing, whereas complete isolation
precludes intermediate interaction.
We therefore introduce a stage-dependent, token-type-aware
mask that controls \emph{when} views interact
and \emph{which} connections are permitted.

\begin{figure}[t]
    \centering
    \includegraphics[width=\linewidth]
    {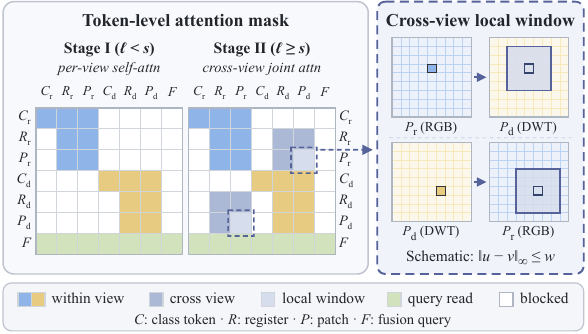}
    \caption{\textbf{Staged joint-attention masks.}
    Rows/columns denote queries/keys. Stage II adds global register
    and local patch connections across views; fusion queries read
    both throughout without writeback. Windows illustrate radius $w$.}
    \label{fig:mask}
\end{figure}

At block $\ell$, let $\mathcal{A}_\ell(t)$ denote the keys visible
to query token $t$. In Eq.~\eqref{eq:joint-attention},
$M^\ell(t,t')$ is zero when key token $t'\in\mathcal{A}_\ell(t)$
and $-\infty$ otherwise. Figure~\ref{fig:mask} shows the connectivity.

\textbf{Stage-dependent connectivity.}
Before block $s$, each view processes its own registers and patches.
From block $s$ onward, registers read all registers and patches of
the opposite view, while patches read its registers and neighboring
patches. Writing $\bar m$ for the opposite view, a query token
$t\in U_m$ has the following visible keys:
\begin{equation}
\mathcal{A}_\ell(t)=U_m\cup
\begin{cases}
    \varnothing,
        & \ell<s,\\
    U_{\bar m},
        & \ell\ge s,\ t\in R_m,\\
    R_{\bar m}\cup
    P_{\bar m}(\mathcal{N}_w(u)),
        & \ell\ge s,\ t=P_m(u).
\end{cases}
\label{eq:staged-attention}
\end{equation}
Here $\mathcal{N}_w(u)$ is the spatial neighborhood defined below.

Class tokens remain view-specific:
\begin{equation}
    \mathcal{A}_\ell(C_m)
    =\{C_m\}\cup U_m,
    \qquad \forall\,\ell.
    \label{eq:class-attention}
\end{equation}
Registers and patches never read class-token keys, but class
tokens can acquire cross-view evidence indirectly through their
updated own-view tokens.

\textbf{Local cross-view patch interaction.}
For aligned patch coordinates $u$ and $v$,
the opposite-view neighborhood is
\begin{equation}
    \mathcal{N}_w(u)
    =\{v\in\Omega:
    \|u-v\|_\infty\leq w\},
    \label{eq:local-neighborhood}
\end{equation}
where $\Omega$ is the common patch grid and $w$ the spatial radius.
This defines a query-centered $(2w+1)\times(2w+1)$ window, clipped
at image boundaries. The window follows each query instead of imposing fixed
single-view partitions. This locality prior restricts direct
patch exchange to spatially corresponding evidence, while global
register connections provide a complementary communication
pathway across views. Within-view attention and fusion-query readout
remain global, unaffected by the spatial window.

\subsection{Progressive fusion-query readout}
\label{sec:method-readout}

Unlike a final-layer linear probe, ReMoT introduces $K$ learnable
fusion queries at the Transformer input and updates them across
all blocks. They read the full token sequence, but visual tokens
cannot read them. Let $\mathcal{T}$ denote the full token set in
Eq.~\eqref{eq:token-sequence}, including both class tokens and all
$K$ fusion queries. At every block,
\begin{equation}
\begin{aligned}
    \mathcal{A}_\ell(F_k)
    &=\mathcal{T},\\
    \mathcal{A}_\ell(t)\cap F
    &=\varnothing,
    \qquad
    t\in\{C_r,C_d\}\cup U_r\cup U_d.
\end{aligned}
\label{eq:readout-mask}
\end{equation}
This one-way connection lets queries collect evidence as the views
evolve, including their later cross-view updates, without affecting
visual-token computation (\emph{read without writeback}). Readout
reuses frozen attention projections without adding a trainable
cross-attention module.

After the last block, the fusion queries
are processed by the frozen final
normalization. Let $f_k$ denote the
normalized representation of query $k$,
and $g$ a shared two-class linear head.
The final prediction is
\begin{equation}
    z_f=\frac{1}{K}\sum_{k=1}^{K}g(f_k),
    \qquad
    p(y=1\mid x)=\operatorname{softmax}(z_f)_1.
    \label{eq:prediction}
\end{equation}
Here, $y=1$ denotes a generated image. We minimize
class-weighted cross-entropy on $z_f$. Only the fusion-query embeddings
and shared classifier are optimized;
all pretrained backbone parameters and
\mbox{input projections} remain frozen.

\begin{table*}[t]
\centering
\caption{Eight-group balanced accuracy (\%). Means $\pm$ sample standard deviations; Avg. weights groups equally. Bold/underline: best/second best. Scoring, sources, and DDA corrections: Appendix~\ref{sec:supp-protocol}.}
\label{tab:main}
\small
\setlength{\tabcolsep}{3pt}
\resizebox{\textwidth}{!}{%
\begin{tabular}{lccccccccc}
\toprule
& \multicolumn{5}{c}{Manually Curated Datasets} & \multicolumn{3}{c}{In-the-Wild Datasets} & \\
\cmidrule(lr){2-6}\cmidrule(lr){7-9}
\raisebox{-.5\height}{Method} & \raisebox{-.5\height}{GI-QF96} & \raisebox{-.5\height}{AIGCB-QF96} & \raisebox{-.5\height}{EvalGEN} & \raisebox{-.5\height}{Synthbuster} & \raisebox{-.5\height}{Mirage-Test} & \raisebox{-.5\height}{Chameleon} & \raisebox{-.5\height}{SynthWildX} & \raisebox{-.5\height}{WildRF} & \raisebox{-.5\height}{Avg.} \\
\midrule
NPR~\cite{ref1} & 51.5 $\pm$ 6.3 & 53.1 $\pm$ 12.2 & 2.9 $\pm$ 2.7 & 50.0 $\pm$ 2.6 & 38.5 $\pm$ 14.0 & 59.9 & 49.8 $\pm$ 10.0 & 63.5 $\pm$ 13.6 & 46.1 $\pm$ 19.0 \\
UnivFD~\cite{ojha2023universal} & 64.1 $\pm$ 10.8 & 72.5 $\pm$ 17.3 & 15.4 $\pm$ 14.2 & 67.8 $\pm$ 14.4 & 25.1 $\pm$ 13.1 & 50.7 & 52.3 $\pm$ 11.3 & 55.3 $\pm$ 5.7 & 50.4 $\pm$ 20.2 \\
FatFormer~\cite{ref0} & 62.8 $\pm$ 10.4 & 85.0 $\pm$ 14.9 & 45.6 $\pm$ 33.1 & 56.1 $\pm$ 10.7 & 36.4 $\pm$ 20.4 & 51.2 & 52.1 $\pm$ 8.2 & 58.9 $\pm$ 8.0 & 56.0 $\pm$ 14.3 \\
SAFE~\cite{li2024safe} & 50.3 $\pm$ 1.2 & 50.3 $\pm$ 1.1 & 1.1 $\pm$ 0.6 & 46.5 $\pm$ 20.8 & 39.8 $\pm$ 22.1 & 59.2 & 49.1 $\pm$ 0.7 & 54.2 $\pm$ 18.5 & 43.8 $\pm$ 18.1 \\
C2P-CLIP~\cite{tan2024c2p} & 74.4 $\pm$ 8.4 & 81.4 $\pm$ 15.6 & 38.9 $\pm$ 31.2 & 68.5 $\pm$ 11.4 & 54.7 $\pm$ 20.1 & 51.1 & 57.1 $\pm$ 4.2 & 59.6 $\pm$ 7.7 & 60.7 $\pm$ 13.6 \\
AIDE~\cite{yan2025sanity} & 61.2 $\pm$ 11.9 & 63.6 $\pm$ 13.9 & 19.1 $\pm$ 11.1 & 53.9 $\pm$ 18.6 & 31.3 $\pm$ 12.1 & 63.1 & 54.7 $\pm$ 7.7 & 58.4 $\pm$ 12.9 & 50.7 $\pm$ 16.4 \\
DRCT~\cite{chen2024drct} & 85.3 $\pm$ 3.9 & 81.4 $\pm$ 12.2 & 77.8 $\pm$ 5.4 & 81.3 $\pm$ 5.8 & 42.0 $\pm$ 17.9 & 56.6 & 55.1 $\pm$ 4.1 & 51.6 $\pm$ 4.5 & 66.4 $\pm$ 16.8 \\
AlignedForensics~\cite{rajan2025aligned} & 79.0 $\pm$ 22.7 & 66.6 $\pm$ 21.6 & 68.0 $\pm$ 20.7 & 77.4 $\pm$ 25.0 & 39.7 $\pm$ 22.1 & 71.0 & 78.8 $\pm$ 17.8 & 80.1 $\pm$ 10.3 & 70.1 $\pm$ 13.4 \\
Effort~\cite{yan2025effort} & 72.7 $\pm$ 15.0 & 71.9 $\pm$ 13.8 & 85.8 $\pm$ 10.1 & 60.5 $\pm$ 11.2 & 43.0 $\pm$ 10.9 & 58.6 & 56.8 $\pm$ 2.8 & 72.2 $\pm$ 6.6 & 65.2 $\pm$ 13.1 \\
DDA~\cite{chen2025dda} & 91.7 $\pm$ 7.8 & \underline{87.8} $\pm$ 12.6 & \textbf{97.2} $\pm$ 4.2 & \underline{90.1} $\pm$ 5.6 & \underline{55.7} $\pm$ 11.2 & \underline{82.4} & \textbf{90.9} $\pm$ 3.1 & \underline{90.3} $\pm$ 3.5 & \underline{85.8} $\pm$ 12.8 \\
OmniAID~\cite{guo2025omniaid} & \textbf{93.4} $\pm$ 6.7 & 85.6 $\pm$ 11.8 & 87.7 $\pm$ 8.3 & 81.9 $\pm$ 8.8 & 51.1 $\pm$ 6.3 & 77.1 & 63.2 $\pm$ 8.7 & 57.9 $\pm$ 5.2 & 74.7 $\pm$ 15.4 \\
\midrule
\rowcolor[gray]{0.94}
\textbf{ReMoT} & \underline{92.3} $\pm$ 7.2 & \textbf{89.1} $\pm$ 9.8 & \underline{96.3} $\pm$ 3.6 & \textbf{93.5} $\pm$ 2.0 & \textbf{84.1} $\pm$ 3.7 & \textbf{90.2} & \underline{90.0} $\pm$ 1.7 & \textbf{94.2} $\pm$ 1.3 & \textbf{91.2} $\pm$ 3.8 \\
\bottomrule
\end{tabular}
}
\end{table*}

\section{Experiments}
\label{sec:experiments}

\begin{table*}[t]
\centering
\caption{Per-generator balanced accuracy on GenImage-QF96 (GI-QF96; \%).}
\label{tab:subset-main}
\small
\setlength{\tabcolsep}{3pt}
\resizebox{0.85\textwidth}{!}{%
\begin{tabular}{lccccccccc}
\toprule
Method & Midjourney & SD~v1.4 & SD~v1.5 & ADM & GLIDE & Wukong & VQDM & BigGAN & Avg. \\
\midrule
NPR~\cite{ref1} & 53.4 & 55.1 & 55.0 & 43.8 & 41.2 & 57.4 & 48.4 & 57.7 & 51.5 $\pm$ 6.3 \\
UnivFD~\cite{ojha2023universal} & 55.1 & 55.6 & 55.7 & 62.5 & 61.3 & 61.1 & 76.9 & 84.4 & 64.1 $\pm$ 10.8 \\
FatFormer~\cite{ref0} & 52.1 & 53.6 & 53.8 & 61.4 & 65.5 & 60.9 & 72.5 & 82.2 & 62.8 $\pm$ 10.4 \\
SAFE~\cite{li2024safe} & 49.0 & 49.7 & 49.8 & 49.5 & 53.0 & 50.3 & 50.2 & 50.9 & 50.3 $\pm$ 1.2 \\
C2P-CLIP~\cite{tan2024c2p} & 56.6 & 77.5 & 76.9 & 71.6 & 73.5 & 79.4 & 73.7 & 85.9 & 74.4 $\pm$ 8.4 \\
AIDE~\cite{yan2025sanity} & 58.2 & 77.2 & 77.4 & 50.4 & 54.6 & 70.5 & 50.8 & 50.6 & 61.2 $\pm$ 11.9 \\
DRCT~\cite{chen2024drct} & 82.4 & 88.3 & 88.2 & 76.9 & 86.1 & 87.9 & 85.4 & 87.0 & 85.3 $\pm$ 3.9 \\
AlignedForensics~\cite{rajan2025aligned} & \textbf{97.5} & \textbf{99.7} & \textbf{99.6} & 52.4 & 57.6 & \textbf{99.6} & 75.0 & 50.6 & 79.0 $\pm$ 22.7 \\
Effort~\cite{yan2025effort} & 72.9 & 91.1 & 90.8 & 56.7 & 75.3 & 80.4 & 64.3 & 49.9 & 72.7 $\pm$ 15.0 \\
DDA~\cite{chen2025dda} & \underline{95.6} & 98.7 & 98.6 & \underline{89.5} & \underline{89.6} & 98.7 & 76.5 & 86.5 & 91.7 $\pm$ 7.8 \\
OmniAID~\cite{guo2025omniaid} & 78.4 & 98.8 & 98.7 & \textbf{92.5} & \textbf{93.2} & 98.6 & \underline{92.4} & \textbf{94.4} & \textbf{93.4} $\pm$ 6.7 \\
\midrule
\rowcolor[gray]{0.94}
\textbf{ReMoT} & 86.1 & \underline{99.0} & \underline{99.0} & 79.2 & 89.2 & \underline{99.0} & \textbf{95.1} & \underline{91.9} & \underline{92.3} $\pm$ 7.2 \\
\bottomrule
\end{tabular}
}
\end{table*}

\begin{table*}[t]
\centering
\caption{Per-generator balanced accuracy on AIGCDetectBench-QF96 (AIGCB-QF96; \%).}
\label{tab:cmp-aigca}
\label{tab:cmp-aigcb}
\small
\setlength{\tabcolsep}{3pt}
\resizebox{\textwidth}{!}{%
\begin{tabular}{lcccccccccccccccccc}
\toprule
Method & \rotatebox{60}{ADM} & \rotatebox{60}{DALL-E2} & \rotatebox{60}{GLIDE} & \rotatebox{60}{Midjourney} & \rotatebox{60}{VQDM} & \rotatebox{60}{BigGAN} & \rotatebox{60}{CycleGAN} & \rotatebox{60}{GauGAN} & \rotatebox{60}{ProGAN} & \rotatebox{60}{SDXL} & \rotatebox{60}{SD~v1.4} & \rotatebox{60}{SD~v1.5} & \rotatebox{60}{StarGAN} & \rotatebox{60}{StyleGAN} & \rotatebox{60}{StyleGAN2} & \rotatebox{60}{WFR} & \rotatebox{60}{Wukong} & Avg. \\
\midrule
NPR~\cite{ref1} & 43.8 & 20.0 & 41.2 & 53.4 & 48.4 & 53.1 & 76.6 & 42.2 & 58.7 & 59.6 & 55.1 & 55.0 & 67.4 & 57.9 & 54.6 & 58.8 & 57.4 & 53.1 $\pm$ 12.2 \\
UnivFD~\cite{ojha2023universal} & 62.5 & 50.0 & 61.3 & 55.1 & 76.9 & 87.5 & \underline{96.9} & \underline{98.8} & \textbf{99.4} & 58.2 & 55.6 & 55.7 & 95.1 & 80.0 & 69.4 & 69.2 & 61.1 & 72.5 $\pm$ 17.3 \\
FatFormer~\cite{ref0} & 80.2 & 68.5 & \underline{91.1} & 54.4 & 88.0 & \textbf{99.2} & \textbf{99.5} & \textbf{99.1} & 98.5 & 71.7 & 67.5 & 67.2 & \underline{99.4} & \textbf{98.0} & \textbf{98.8} & \underline{88.3} & 75.6 & 85.0 $\pm$ 14.9 \\
SAFE~\cite{li2024safe} & 49.5 & 49.5 & 53.0 & 49.0 & 50.2 & 52.2 & 51.9 & 50.0 & 50.0 & 49.8 & 49.7 & 49.8 & 50.1 & 50.0 & 50.0 & 49.8 & 50.3 & 50.3 $\pm$ 1.1 \\
C2P-CLIP~\cite{tan2024c2p} & 71.6 & 52.3 & 73.5 & 56.6 & 73.7 & \underline{98.4} & 96.8 & \underline{98.8} & \underline{99.3} & 62.3 & 77.5 & 76.9 & \textbf{99.6} & \underline{93.1} & 79.4 & \textbf{94.8} & 79.4 & 81.4 $\pm$ 15.6 \\
AIDE~\cite{yan2025sanity} & 52.9 & 51.1 & 60.2 & 49.8 & 69.3 & 70.1 & 93.6 & 60.6 & 89.0 & 49.6 & 51.6 & 51.0 & 72.1 & 66.5 & 59.0 & 80.6 & 54.5 & 63.6 $\pm$ 13.9 \\
DRCT~\cite{chen2024drct} & 79.9 & \underline{89.2} & 89.2 & 85.5 & 88.6 & 81.4 & 91.0 & 93.8 & 71.1 & 88.3 & 91.4 & 91.0 & 53.0 & 62.7 & 63.8 & 73.9 & 90.8 & 81.4 $\pm$ 12.2 \\
AlignedForensics~\cite{rajan2025aligned} & 51.6 & 52.0 & 55.6 & \textbf{96.2} & 72.1 & 51.2 & 49.5 & 50.8 & 50.7 & 95.1 & \textbf{99.7} & \textbf{99.6} & 53.8 & 52.7 & 51.6 & 50.0 & \textbf{99.6} & 66.6 $\pm$ 21.6 \\
Effort~\cite{yan2025effort} & 56.7 & 71.6 & 75.3 & 72.9 & 64.3 & 69.7 & 74.8 & 54.7 & 69.5 & 81.1 & 91.1 & 90.8 & 96.1 & 52.7 & 47.8 & 73.7 & 80.4 & 71.9 $\pm$ 13.8 \\
DDA~\cite{chen2025dda} & \underline{89.5} & \textbf{94.6} & 89.6 & \underline{95.6} & 76.6 & 91.0 & 72.5 & 92.7 & 92.8 & \textbf{99.4} & 98.7 & 98.6 & 72.7 & 87.8 & \underline{90.2} & 52.1 & 98.8 & \underline{87.8} $\pm$ 12.6 \\
OmniAID~\cite{guo2025omniaid} & \textbf{92.5} & 79.6 & \textbf{93.2} & 78.4 & \underline{92.4} & 86.6 & 92.8 & 86.5 & 56.6 & 86.9 & 98.8 & 98.7 & 88.4 & 74.4 & 64.9 & 86.3 & 98.6 & 85.6 $\pm$ 11.8 \\
\midrule
\rowcolor[gray]{0.94}
\textbf{ReMoT} & 79.2 & 83.7 & 89.2 & 86.1 & \textbf{95.1} & 97.1 & 83.2 & 97.6 & 94.1 & \underline{98.3} & \underline{99.0} & \underline{99.0} & 63.1 & 84.5 & 84.4 & 81.5 & \underline{99.0} & \textbf{89.1} $\pm$ 9.8 \\
\bottomrule
\end{tabular}
}
\end{table*}
\begin{table*}[t]
\centering
\caption{Mirage-Test (\%): class-wise accuracy and overall balanced accuracy. Anime reports fake-image accuracy.}
\label{tab:cmp-mirage}
\small
\setlength{\tabcolsep}{2.5pt}
\resizebox{\textwidth}{!}{%
\begin{tabular}{lcccccccccccccccc}
\toprule
& \multicolumn{3}{c}{Human} & \multicolumn{3}{c}{Animal} & \multicolumn{3}{c}{Object} & \multicolumn{3}{c}{Scene} & \multicolumn{3}{c}{Anime} & \\
\cmidrule(lr){2-4}\cmidrule(lr){5-7}\cmidrule(lr){8-10}\cmidrule(lr){11-13}\cmidrule(lr){14-16}
Method & Real & Fake & Overall & Real & Fake & Overall & Real & Fake & Overall & Real & Fake & Overall & Real & Fake & Overall & Mean \\
\midrule
NPR & 79.32 & 12.17 & 45.74 & 68.86 & 17.91 & 43.39 & 77.12 & 12.67 & 44.89 & 71.92 & 18.00 & 44.96 & -- & 13.45 & 13.45 & 38.49 \\
UnivFD & 67.32 & 1.20 & 34.26 & 45.57 & 3.51 & 24.54 & 55.13 & 7.88 & 31.51 & 61.12 & 4.40 & 32.76 & -- & 2.58 & 2.58 & 25.13 \\
FatFormer & 90.35 & 0.03 & 45.19 & 87.94 & 0.00 & 43.97 & 94.62 & 0.00 & 47.31 & 90.60 & 0.02 & 45.31 & -- & 0.02 & 0.02 & 36.36 \\
SAFE & \underline{99.42} & 0.02 & 49.72 & \underline{99.20} & 0.17 & 49.69 & \textbf{98.73} & 0.37 & 49.55 & \textbf{99.38} & 0.03 & 49.71 & -- & 0.33 & 0.33 & 39.80 \\
C2P-CLIP & 7.25 & \underline{85.43} & 46.34 & 9.37 & \underline{82.17} & 45.77 & 10.22 & \underline{77.88} & 44.05 & 13.18 & \underline{80.50} & 46.84 & -- & \textbf{90.55} & \textbf{90.55} & 54.71 \\
AIDE & 62.93 & 10.02 & 36.48 & 61.94 & 15.37 & 38.66 & 54.97 & 10.00 & 32.49 & 67.28 & 10.00 & 38.64 & -- & 10.02 & 10.02 & 31.25 \\
DRCT & 90.20 & 6.12 & 48.16 & 93.43 & 13.77 & 53.60 & 91.58 & 7.17 & 49.38 & 91.97 & 5.63 & 48.80 & -- & 10.28 & 10.28 & 42.04 \\
AlignedForensics & \textbf{99.43} & 0.12 & 49.77 & \textbf{99.37} & 0.14 & 49.76 & \underline{98.27} & 0.13 & 49.20 & \underline{99.30} & 0.08 & 49.69 & -- & 0.12 & 0.12 & 39.71 \\
Effort & 67.98 & 24.07 & 46.03 & 81.26 & 27.57 & 54.41 & 57.72 & 21.38 & 39.55 & 64.90 & 33.12 & 49.01 & -- & 26.13 & 26.13 & 43.03 \\
DDA & 92.50 & 38.72 & \underline{65.61} & 89.34 & 37.34 & \underline{63.34} & 93.12 & 22.12 & \underline{57.62} & 90.18 & 18.98 & \underline{54.58} & -- & 37.22 & 37.22 & \underline{55.67} \\
OmniAID & 76.40 & 42.35 & 59.38 & 82.63 & 29.17 & 55.90 & 82.60 & 15.43 & 49.02 & 80.60 & 12.45 & 46.53 & -- & 44.67 & 44.67 & 51.10 \\
\midrule
\rowcolor[gray]{0.94}
\textbf{ReMoT} & 86.88 & \textbf{92.63} & \textbf{89.76} & 82.57 & \textbf{83.89} & \textbf{83.23} & 80.83 & \textbf{78.15} & \textbf{79.49} & 83.18 & \textbf{83.32} & \textbf{83.25} & -- & \underline{84.83} & \underline{84.83} & \textbf{84.11} \\
\bottomrule
\end{tabular}
}
\end{table*}

\begin{table*}[t]
\centering
\caption{Balanced accuracy on SynthWildX and WildRF (\%).}
\label{tab:cmp-synx}
\label{tab:cmp-wildrf}
\small
\setlength{\tabcolsep}{3pt}
\begin{tabular}{lcccc|cccc}
\toprule
& \multicolumn{4}{c|}{SynthWildX} & \multicolumn{4}{c}{WildRF} \\
\cmidrule(lr){2-5}\cmidrule(lr){6-9}
Method & DALL-E3 & Firefly & Midjourney & Avg. & Facebook & Reddit & Twitter & Avg. \\
\midrule
NPR~\cite{ref1} & 43.6 & 61.3 & 44.5 & 49.8 $\pm$ 10.0 & 78.1 & 61.0 & 51.3 & 63.5 $\pm$ 13.6 \\
UnivFD~\cite{ojha2023universal} & 45.4 & 65.3 & 46.2 & 52.3 $\pm$ 11.3 & 49.1 & 60.2 & 56.5 & 55.3 $\pm$ 5.7 \\
FatFormer~\cite{ref0} & 46.5 & 61.6 & 48.3 & 52.1 $\pm$ 8.2 & 54.1 & 68.1 & 54.4 & 58.9 $\pm$ 8.0 \\
SAFE~\cite{li2024safe} & 49.4 & 48.2 & 49.6 & 49.1 $\pm$ 0.7 & 50.9 & 74.1 & 37.5 & 54.2 $\pm$ 18.5 \\
C2P-CLIP~\cite{tan2024c2p} & 56.9 & 61.4 & 53.0 & 57.1 $\pm$ 4.2 & 54.4 & 68.4 & 55.9 & 59.6 $\pm$ 7.7 \\
AIDE~\cite{yan2025sanity} & 63.4 & 48.8 & 51.9 & 54.7 $\pm$ 7.7 & 57.8 & 71.5 & 45.8 & 58.4 $\pm$ 12.9 \\
DRCT~\cite{chen2024drct} & 58.3 & 56.4 & 50.5 & 55.1 $\pm$ 4.1 & 46.6 & 53.1 & 55.2 & 51.6 $\pm$ 4.5 \\
AlignedForensics~\cite{rajan2025aligned} & 85.5 & 58.5 & \underline{92.2} & 78.8 $\pm$ 17.8 & 89.4 & 69.1 & 81.8 & 80.1 $\pm$ 10.3 \\
Effort~\cite{yan2025effort} & 55.1 & 60.0 & 55.3 & 56.8 $\pm$ 2.8 & 70.6 & 79.4 & 66.6 & 72.2 $\pm$ 6.6 \\
DDA~\cite{chen2025dda} & \textbf{92.3} & \underline{87.3} & \textbf{93.1} & \textbf{90.9} $\pm$ 3.1 & \underline{93.1} & \underline{86.4} & \underline{91.5} & \underline{90.3} $\pm$ 3.5 \\
OmniAID~\cite{guo2025omniaid} & 71.7 & 63.6 & 54.4 & 63.2 $\pm$ 8.7 & 54.1 & 63.9 & 55.9 & 57.9 $\pm$ 5.2 \\
\midrule
\rowcolor[gray]{0.94}
\textbf{ReMoT} & \underline{90.9} & \textbf{88.1} & 91.1 & \underline{90.0} $\pm$ 1.7 & \textbf{95.6} & \textbf{93.2} & \textbf{93.8} & \textbf{94.2} $\pm$ 1.3 \\
\bottomrule
\end{tabular}
\end{table*}
\begin{table*}[t]
\centering
\caption{Component ablations in balanced accuracy (\%; scoring as in Table~\ref{tab:main}). Avg. weights groups equally. Mask variants change only cross-view connections.}
\label{tab:ablation}
\small
\resizebox{0.9\textwidth}{!}{%
\begin{tabular}{lccccccccc}
\toprule
Configuration & GI-QF96 & AIGCB-QF96 & EvalGEN & SynB & Mirage & Cham & SynX & WildRF & Avg. \\
\midrule
\multicolumn{10}{l}{\textit{Readout and adaptation}} \\
DINOv3 Linear Probe & 87.7 & 83.3 & 96.5 & 90.1 & 62.9 & 83.1 & 85.8 & 84.0 & 84.2 \\
ReMoT + attention LoRA ($r=8$) & 83.8 & 72.4 & 77.3 & 73.1 & 47.0 & 57.4 & 78.0 & 75.5 & 70.5 \\
\midrule
\multicolumn{10}{l}{\textit{Input views}} \\
DWT-only query readout & 63.6 & 57.1 & 86.7 & 65.7 & 69.5 & 67.5 & 64.0 & 70.0 & 68.0 \\
RGB-only query readout & 91.1 & 86.5 & \textbf{98.0} & 91.5 & 73.6 & 80.9 & 85.5 & 85.6 & 86.6 \\
\midrule
\multicolumn{10}{l}{\textit{Cross-view attention}} \\
No cross-view attention ($s=24$) & 89.4 & 84.3 & \textbf{98.0} & 89.3 & 80.5 & 88.6 & 83.0 & 82.9 & 87.0 \\
Local patch attention only & \textbf{92.4} & 88.8 & 97.1 & 92.0 & 83.0 & 87.2 & 85.9 & 90.3 & 89.6 \\
Global register attention only & 89.0 & 84.0 & 89.8 & 93.3 & 82.0 & 89.9 & 89.2 & \textbf{95.6} & 89.1 \\
Global patch + global register attention & 91.8 & 88.6 & 96.2 & 92.1 & 83.6 & 89.9 & 88.5 & 93.4 & 90.5 \\
\midrule
\rowcolor[gray]{0.94}
\textbf{ReMoT} & 92.3 & \textbf{89.1} & 96.3 & \textbf{93.5} & \textbf{84.1} & \textbf{90.2} & \textbf{90.0} & 94.2 & \textbf{91.2} \\
\bottomrule
\end{tabular}}
\end{table*}
\begin{figure*}[t!]
\centering
\includegraphics[width=\textwidth]{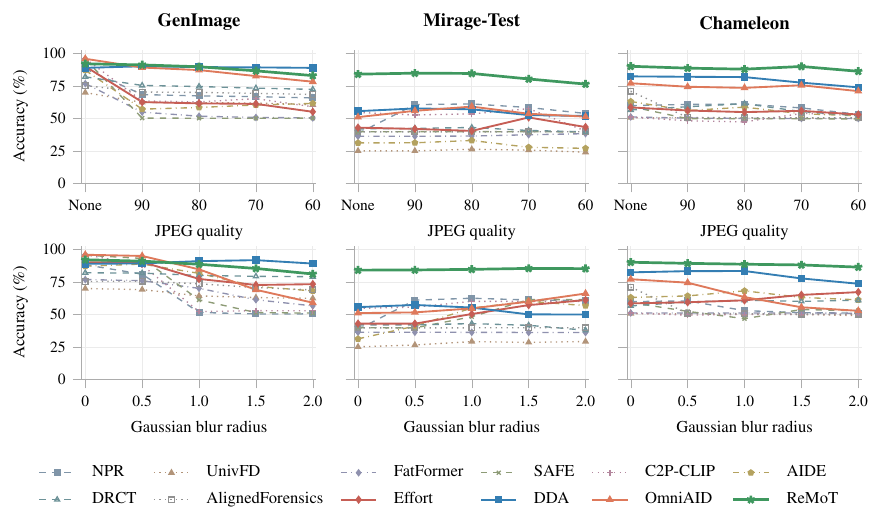}
\caption{Robustness to JPEG compression (top) and Gaussian blur
(bottom). ``None'' and radius 0 denote baseline conditions.
Selected Mirage-Test/Chameleon baseline markers use Table~\ref{tab:main}
values; all other points are measured locally. GenImage uses original
validation images, not GI-QF96. Sources and protocol:
Appendix~\ref{sec:supp-robustness}.}
\label{fig:robustness}
\end{figure*}

\subsection{Experimental setup}
\label{sec:experimental-setup}
\paragraph{Implementation details.}
We train ReMoT on a single NVIDIA H200 GPU using only the
GenImage SD~v1.4 training partition for ten epochs, including
one warmup epoch. The model uses frozen DINOv3 ViT-L/16,
$336\times336$ inputs, one-level db2 DWT, two fusion queries,
a stage boundary of $s=9$, and cross-view patch radius $w=5$.
The readout is optimized with class-weighted
cross-entropy, batch size 96, NAdam, learning rate $7.5\times10^{-5}$,
and weight decay 0.05. We apply Gaussian blur, JPEG compression,
resizing, and noise augmentations. The exact equivalent implementation
uses \textbf{3,074 trainable parameters}
(Appendix~\ref{sec:supp-equivalence}); preprocessing is detailed in
Appendix~\ref{sec:supp-protocol}.
\emph{All ReMoT evaluations use the same trained model, without
dataset-specific fine-tuning or threshold adjustments.}

\paragraph{Evaluation benchmarks and metrics.}
We evaluate across eight benchmark groups:
GenImage (GI)~\cite{zhu2023genimage},
AIGCDetectBench (AIGCB)~\cite{zhong2023patchcraft},
EvalGEN~\cite{chen2025dda},
Synthbuster (SynB)~\cite{bammey2023synthbuster},
Mirage-Test (Mir)~\cite{guo2025omniaid},
Chameleon (Cham)~\cite{yan2025sanity},
SynthWildX (SynX)~\cite{cozzolino2024clip},
and WildRF~\cite{cavia2024wildrf}.
Following DDA~\cite{chen2025dda}, we reduce format bias by JPEG-compressing
synthetic images in GenImage and AIGCDetectBench at quality 96
(GI-QF96 and AIGCB-QF96).
We report balanced accuracy to account for class imbalance
(accuracy for single-class subsets), averaging subsets within groups
and weighting all eight groups equally.

\paragraph{Comparative methods.}
We compare ReMoT with NPR~\cite{ref1},
UnivFD~\cite{ojha2023universal}, FatFormer~\cite{ref0},
SAFE~\cite{li2024safe}, C2P-CLIP~\cite{tan2024c2p},
AIDE~\cite{yan2025sanity}, DRCT~\cite{chen2024drct},
AlignedForensics~\cite{rajan2025aligned}, Effort~\cite{yan2025effort},
DDA~\cite{chen2025dda}, and OmniAID~\cite{guo2025omniaid}.

\subsection{Main results}
\label{sec:comparisons}
\textbf{Broad cross-benchmark transfer.} ReMoT ranks first on five
groups and second on three, averaging 91.2\% (Table~\ref{tab:main}). Mirage-Test exposes severe transfer
failures among baselines, with even the strongest reaching only
55.7\%, versus ReMoT's 84.1\%. The lead also holds across the other
seven groups: ReMoT averages approximately 92.2\%, ahead of DDA's
90.1\%. Transfer also spans unseen generators in GI-QF96/AIGCB-QF96
(Tables~\ref{tab:subset-main} and~\ref{tab:cmp-aigcb}) and nonphotographic
Anime images in Mirage-Test (84.83\%). Together, these results show
that ReMoT's overall advantage is sustained across diverse benchmarks
rather than driven by an isolated gain on a single dataset.

\noindent\textbf{Consistency and class balance.} ReMoT combines high accuracy with the lowest eight-group
standard deviation (3.8 points), indicating consistent transfer. Mirage-Test further reveals
class-prediction bias (Table~\ref{tab:cmp-mirage}): SAFE and
AlignedForensics classify almost all photographic inputs as real,
whereas C2P-CLIP strongly favors fake predictions. ReMoT has the
smallest real--fake accuracy gap in all four photographic categories,
with both class accuracies above 78\%.
This balanced operating point combines substantially higher
fake-image recall than DDA with real-image accuracy above 80\%
in every photographic category. On Synthbuster,
a higher mean also accompanies lower subset variability
(Appendix Table~\ref{tab:cmp-synthbuster}). These patterns support
transfer beyond narrow training-generator cues.

\noindent\textbf{In-the-wild generalization.} Deployment requires
generalization to diverse, uncontrolled image sources.
Trained solely on SD~v1.4, ReMoT is the only compared
detector reaching at least 90\% on all three in-the-wild benchmarks
(Table~\ref{tab:main}). Knowledge readout thus enables transfer from
a single training generator to real-world images, supporting deployment
without source-specific adaptation.

\subsection{Ablation studies}
\label{sec:ablations}
The ablations support learning to read frozen features and combining
complementary views through selective interaction
(Table~\ref{tab:ablation}). ReMoT reaches 91.2\% average accuracy,
compared with 84.2\% for final-layer RGB linear probing.
Adding modality-specific attention LoRA ($r=8$, $\alpha=16$)
increases trainable parameters to approximately 2.37M but lowers
accuracy to 70.5\%. This comparison supports learning to read
pretrained features rather than learning task-specific feature transformations.
For the input views, RGB-only and DWT-only
query readouts achieve 86.6\% and 68.0\%: RGB provides the stronger
individual evidence, while combining both yields the best average.
Retaining fusion-query readout while disabling cross-view attention
($s=24$) yields 87.0\%,
4.2 points below ReMoT.
ReMoT improves seven of eight groups, most notably SynthWildX
(+7.1 points) and WildRF (+11.3); EvalGEN scores are 96.3\%
and 98.0\%, respectively.
The mask variants then examine how views should communicate:
local patch attention alone and global register attention alone
reach 89.6\% and 89.1\%, versus 91.2\% with both. Both ablations
preserve within-view attention and query readout, supporting complementary local
correspondence and global communication. Making patch attention
global instead yields 90.5\%, with lower scores on all eight groups
despite retaining global register communication. These results
favor spatially selective patch exchange alongside global register
communication. Among mask variants, local-only leads on GI-QF96 and EvalGEN, and
register-only on WildRF; combining both pathways provides the
strongest overall transfer. Additional parameter
studies appear in Appendix~\ref{sec:supp-parameters}.

\subsection{Robustness to image post-processing}
\label{sec:robustness}
ReMoT combines high post-processing accuracy with strong performance
retention (Figure~\ref{fig:robustness}).
On Chameleon, ReMoT retains 86.3\% accuracy under JPEG quality 60
and 86.4\% at blur radius 2.0, preserving 95.7\% and 95.8\%
of its unmodified accuracy. DDA retains 89.8\% and 89.4\%,
respectively, so ReMoT's relative accuracy loss is less than half
as large. At these strongest tested severities, its margin over
DDA widens from 7.8 points on unmodified images to 12.3 and
12.8 points. The gain thus reflects both a stronger starting
point and better preservation of detection performance.
On original GenImage, ReMoT obtains 83.0\% under JPEG quality 60
and 81.2\% at blur radius 2.0; DDA reaches 89.0\% and 89.2\%.
On Mirage-Test, ReMoT leads all compared methods under all eight
degradation settings by at least 19.1 points. JPEG quality 60
yields 76.5\%, retaining 90.9\% of its unmodified accuracy,
while blur radius 2.0 yields 85.3\%, exceeding its 84.1\%
baseline.

\FloatBarrier
\section{Conclusion}
We proposed ReMoT to address transfer risks from backbone adaptation
and the limited scope of existing readout approaches.
It couples staged cross-view interaction with progressive evidence
collection across Transformer depth, while keeping pretrained
backbone weights frozen. Trained on a single generator, ReMoT
achieves broad cross-benchmark generalization,
balanced predictions, and strong robustness
to image post-processing. These findings motivate a research focus
on readout design: organizing complementary evidence and learning
how to access pretrained knowledge provide an effective basis for
generalization across unfamiliar generators and diverse image sources.

\section*{LLM Usage Statement}
OpenAI Codex assisted with manuscript preparation,
including language refinement and the presentation of figures and tables.
The authors were responsible for the research ideas, methodological design,
experimental implementation and evaluation, and interpretation of the results.
The authors take full responsibility for the scientific content,
experimental results, and final submission.

{
    \small
    \bibliographystyle{ieeenat_fullname}
    \bibliography{main}
}
\clearpage
\maketitlesupplementary

\appendix
\section{\texorpdfstring{Additional implementation\\and reproducibility details}{Additional implementation and reproducibility details}}
\label{sec:supp-protocol}

\paragraph{Baseline sources.}
We use published baseline scores where available and otherwise
evaluate the authors' released checkpoints. We obtain most published
scores from DDA~\cite{chen2025dda} and published Mirage-Test scores
from OmniAID~\cite{guo2025omniaid}. Where DDA's reported benchmark
averages differ from the arithmetic means of its listed subset scores,
we retain the subset scores and recompute the corresponding averages.
GI-QF96 and AIGCB-QF96 apply JPEG quality 96 to
synthetic images in GenImage and AIGCDetectBench, respectively.

\paragraph{Evaluation metrics.}
For subsets containing both classes, balanced accuracy is the
mean of real-image and fake-image classification accuracy.
For fake-only subsets, including EvalGEN and the Anime subset
of Mirage-Test, we report fake-image classification accuracy,
since real-image accuracy cannot be computed without real samples.
All subsets use the same fixed decision threshold of $0.5$.
Subset scores are averaged within each benchmark group, and
the overall average assigns equal weight to all eight group means.

\paragraph{Wavelet embedding and augmentation.}
The frozen DINOv3 ViT-L/16 has 24 blocks of hidden width 1024 and
four register tokens per view. DWT is computed on the augmented
RGB image in $[0,1]$, before RGB normalization. For a
$336\times336$ input, zero-padded db2 subbands are cropped to
$168\times168$ by retaining their first 168 rows and columns. The frozen 12-channel
DWT patch stem is obtained by subsampling the pretrained RGB
$16\times16$ kernel into an $8\times8$ kernel, repeating its input
channel weights over four wavelet subbands, and scaling both the
weights and bias by $1/4$. The two aligned patch grids use matching
2D RoPE coordinates. Gaussian blur uses $\sigma\in[0,3]$;
JPEG compression, resizing, and noise have application probabilities
of 0.15, 0.5, and 0.1, respectively, with randomized augmentation
order. During evaluation, all preprocessing operations are deterministic.

\paragraph{Query initialization and ablation provenance.}
The two logical fusion queries are initialized from the same
pretrained class embedding; the equivalent implementation stores
one query with a fixed fusion-key bias. The DINOv3 linear probe trains a classification head
on frozen RGB ViT-L/16 class-token features using the GenImage
SD~v1.4 training partition. The single-view query readouts use either RGB or DWT inputs.
The no-interaction model retains both views and the same
fusion-query pathway, with $s=24$ disabling direct
cross-view token attention in all blocks.

\section{Additional generator-level results}
\label{sec:supp-detailed-results}

Tables~\ref{tab:cmp-synthbuster} and~\ref{tab:cmp-evalgen}
provide the per-generator results on Synthbuster and EvalGEN,
respectively, showing how performance varies across generators
within each benchmark.

\begin{table*}[t]
\centering
\caption{Per-generator balanced accuracy on Synthbuster (\%).}
\label{tab:cmp-synthbuster}
\small
\setlength{\tabcolsep}{3pt}
\resizebox{0.85\textwidth}{!}{%
\begin{tabular}{lcccccccccc}
\toprule
Method & DALL-E2 & DALL-E3 & Firefly & GLIDE & Midjourney & SD~v1.3 & SD~v1.4 & SD~v2 & SDXL & Avg. \\
\midrule
NPR~\cite{ref1} & 51.1 & 49.3 & 46.5 & 48.5 & 52.8 & 51.4 & 51.8 & 46.0 & 52.8 & 50.0 $\pm$ 2.6 \\
UnivFD~\cite{ojha2023universal} & 83.5 & 47.4 & \underline{89.9} & 53.3 & 52.5 & 70.4 & 69.9 & 75.7 & 68.0 & 67.8 $\pm$ 14.4 \\
FatFormer~\cite{ref0} & 59.4 & 39.5 & 60.3 & 72.7 & 44.4 & 53.7 & 54.0 & 52.3 & 69.1 & 56.1 $\pm$ 10.7 \\
SAFE~\cite{li2024safe} & 58.0 & 9.9 & 10.3 & 52.2 & 56.7 & 59.4 & 59.1 & 53.0 & 59.5 & 46.5 $\pm$ 20.8 \\
C2P-CLIP~\cite{tan2024c2p} & 55.6 & 63.2 & 59.5 & 86.7 & 52.9 & 75.2 & 76.7 & 69.2 & 77.7 & 68.5 $\pm$ 11.4 \\
AIDE~\cite{yan2025sanity} & 34.9 & 33.7 & 24.8 & 65.0 & 57.5 & 74.1 & 73.7 & 53.2 & 68.4 & 53.9 $\pm$ 18.6 \\
DRCT~\cite{chen2024drct} & 77.2 & 86.6 & 84.1 & 82.6 & 73.7 & 86.6 & 86.6 & 83.2 & 71.3 & 81.3 $\pm$ 5.8 \\
AlignedForensics~\cite{rajan2025aligned} & 50.2 & 48.9 & 51.7 & 53.5 & \textbf{98.7} & \textbf{98.8} & \textbf{98.8} & \textbf{98.6} & \textbf{97.3} & 77.4 $\pm$ 25.0 \\
Effort~\cite{yan2025effort} & 51.1 & 62.5 & 35.0 & 66.6 & 59.1 & 69.8 & 69.4 & 63.2 & 67.7 & 60.5 $\pm$ 11.2 \\
DDA~\cite{chen2025dda} & 86.3 & 90.0 & \textbf{91.9} & 76.5 & \underline{93.5} & 92.9 & 92.7 & 93.3 & 93.5 & \underline{90.1} $\pm$ 5.6 \\
OmniAID~\cite{guo2025omniaid} & 75.0 & 87.4 & 71.4 & \underline{89.0} & 66.1 & 89.6 & 89.5 & 84.0 & 85.6 & 81.9 $\pm$ 8.8 \\
DINOv3 Linear Probe & \underline{88.7} & \underline{91.8} & 81.3 & 86.4 & 91.1 & 93.0 & 93.2 & 92.6 & 93.1 & \underline{90.1} $\pm$ 4.0 \\
\midrule
\rowcolor[gray]{0.94}
\textbf{ReMoT} & \textbf{92.6} & \textbf{95.2} & 89.8 & \textbf{91.3} & 93.3 & \underline{95.1} & \underline{95.1} & \underline{94.5} & \underline{95.1} & \textbf{93.5} $\pm$ 2.0 \\
\bottomrule
\end{tabular}
}
\end{table*}

\begin{table*}[t]
\centering
\caption{Per-generator fake-image detection accuracy on EvalGEN (\%).}
\label{tab:cmp-evalgen}
\small
\begin{tabular}{lcccccc}
\toprule
Method & Flux & GoT & Infinity & NOVA & OmniGen & Avg. \\
\midrule
NPR~\cite{ref1} & 0.7 & 0.2 & 6.5 & 4.7 & 2.2 & 2.9 $\pm$ 2.7 \\
UnivFD~\cite{ojha2023universal} & 4.0 & 9.2 & 15.7 & 8.3 & 39.6 & 15.4 $\pm$ 14.2 \\
FatFormer~\cite{ref0} & 9.9 & 47.9 & 44.7 & 98.3 & 27.3 & 45.6 $\pm$ 33.1 \\
SAFE~\cite{li2024safe} & 1.0 & 0.5 & 1.9 & 0.6 & 1.6 & 1.1 $\pm$ 0.6 \\
C2P-CLIP~\cite{tan2024c2p} & 8.7 & 49.6 & 35.3 & 86.4 & 14.5 & 38.9 $\pm$ 31.2 \\
AIDE~\cite{yan2025sanity} & 17.9 & 24.7 & 3.4 & 16.3 & 33.4 & 19.1 $\pm$ 11.1 \\
DRCT~\cite{chen2024drct} & 72.5 & 81.4 & 77.9 & 84.6 & 72.5 & 77.8 $\pm$ 5.4 \\
AlignedForensics~\cite{rajan2025aligned} & 32.0 & 72.3 & 74.0 & 84.8 & 77.0 & 68.0 $\pm$ 20.7 \\
Effort~\cite{yan2025effort} & 72.2 & 96.7 & 80.5 & 85.4 & 94.2 & 85.8 $\pm$ 10.1 \\
DDA~\cite{chen2025dda} & 89.9 & \textbf{99.5} & \textbf{97.8} & \underline{99.5} & \underline{99.5} & \textbf{97.2} $\pm$ 4.2 \\
OmniAID~\cite{guo2025omniaid} & 75.9 & 96.4 & 82.8 & 93.7 & 89.5 & 87.7 $\pm$ 8.3 \\
DINOv3 Linear Probe & \textbf{93.9} & \underline{98.9} & 89.5 & \textbf{100.0} & \textbf{100.0} & \underline{96.5} $\pm$ 4.6 \\
\midrule
\rowcolor[gray]{0.94}
\textbf{ReMoT} & \underline{91.1} & 97.8 & \underline{93.9} & 99.3 & 99.3 & 96.3 $\pm$ 3.6 \\
\bottomrule
\end{tabular}
\end{table*}

\section{Exact reparameterizations and efficiency}
\label{sec:supp-equivalence}

\paragraph{Design and equivalent implementation.}
ReMoT's MoT-inspired formulation assigns separate Transformer
parameter slots to the RGB and DWT streams and uses $K$ fusion
queries for progressive readout. Under our training setup, we find
that ReMoT admits a more compact, exactly equivalent implementation.
It shares the slot weights and replaces the $K$ queries with one
query and a fixed attention bias, preserving both visual streams,
their staged interaction, and the model's predictions.
Weight sharing relies on identical slot
parameters, which we maintain by freezing both slots at the same
pretrained initialization. Query reduction relies on identical
initial query states and symmetric computation at every layer.
Modality-specific updates that make the slot weights differ or
query-specific changes that break symmetry invalidate the
respective equivalence.

\paragraph{Frozen Transformer-slot sharing.}
The MoT-inspired formulation assigns RGB and DWT tokens to
view-specific Transformer parameter slots. In ReMoT, however,
both slots are initialized from the same pretrained DINOv3
checkpoint and remain frozen. Denote all parameters of block
$\ell$ in view $m$ by $\Theta_{m,\ell}$, including normalization,
QKV and output projections, and the feed-forward network. Then
$\Theta_{r,\ell}=\Theta_{d,\ell}=\Theta_\ell$ for every block.
Replacing both view-specific parameter slots with the shared
parameters $\Theta_\ell$ therefore
preserves the projections of \emph{every} token:
\begin{equation}
    (\mathbf{Q}^\ell,\mathbf{K}^\ell,\mathbf{V}^\ell)_{\mathrm{MoT}}
    =(\mathbf{Q}^\ell,\mathbf{K}^\ell,\mathbf{V}^\ell)_{\mathrm{shared}}.
    \label{eq:supp-slot-projection}
\end{equation}
Both realizations apply the same spatial RoPE rotations and
stage-dependent mask $M^\ell$ to the full concatenated sequence.
Their joint-attention outputs consequently agree, as do the
subsequent output projections, residual updates, normalization,
and feed-forward operations. Writing $\mathcal B_\ell$ for the
complete masked Transformer block yields
\begin{equation}
\begin{aligned}
Z_{\mathrm{MoT}}^{\ell+1}
&=\mathcal B_\ell(Z^\ell;\Theta_{r,\ell},
                       \Theta_{d,\ell},M^\ell)\\
&=\mathcal B_\ell(Z^\ell;\Theta_\ell,M^\ell)
 =Z_{\mathrm{shared}}^{\ell+1}.
\end{aligned}
\label{eq:supp-slot-block}
\end{equation}
Since both realizations start from the same RGB, DWT, and
fusion-token sequence, induction over Transformer depth proves
identical final representations and predictions. Importantly,
this shares \emph{parameters}, not view tokens: the two input
stems, all view-token computations, and all stage-dependent
cross-view connections remain intact. Equality of the two slot
parameter sets is the sufficient condition for this reduction;
freezing ensures that the two sets remain equal throughout training.
Modality-specific updates that make the slots differ prevent exact
sharing of the affected parameters. Unfreezing parameters that
remain tied preserves their equality.

\paragraph{Symmetric fusion-query reduction.}
The $K$ fusion queries share the same initialization, projections,
attention visibility, and prediction head, without query-specific
positional information. Every Transformer block is therefore
equivariant to permutations of fusion-query indices. Identical
initial queries remain identical at every depth,
$f_1^\ell=\cdots=f_K^\ell$. For one fusion-query attention head,
let $\mathcal V$ denote the visible non-fusion keys and let
$(s_f,v_f)$ denote the common fusion-key logit and value. Its
output is
\begin{equation}
    o_f^{(K)}=
    \frac{\sum_{j\in\mathcal V}e^{s_j}v_j
          +K e^{s_f}v_f}
         {\sum_{j\in\mathcal V}e^{s_j}+K e^{s_f}}.
    \label{eq:supp-query-reduction}
\end{equation}
Because $K e^{s_f}=e^{s_f+\log K}$, retaining one fusion query
and adding $\log K$ to its fusion-key attention logit reproduces
the original readout exactly. ReMoT's no-writeback mask also
leaves visual-token updates unchanged, establishing the equivalence
across all blocks. This symmetry depends on the query computation,
not on whether \mbox{backbone transformations} are frozen.
Training shared projections preserves the reduction as long as
query states and computations remain symmetric. Different query
initializations, positional encodings, or visibility rules can
break this symmetry.

\begin{figure*}[t]
\centering
\includegraphics[width=0.98\textwidth]{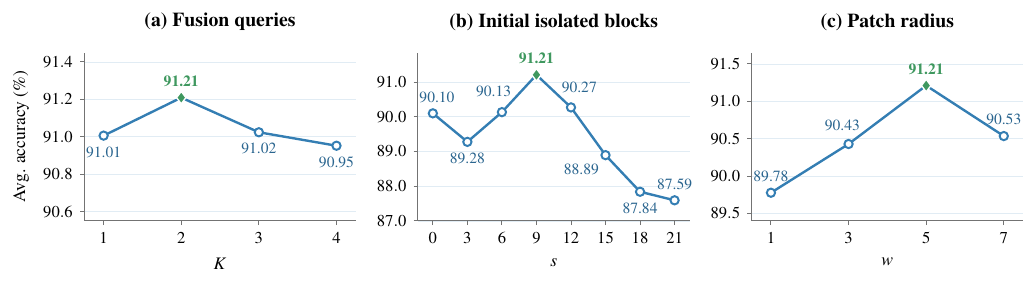}
\caption{Parameter exploration across eight benchmark groups.
Lines connect evaluated settings, and diamonds mark the highest score in each panel.
Panels use different accuracy ranges, with scores shown to two decimal places.
In (b), $s$ counts the initial blocks in which the views are isolated.
Direct exchange starts in the first block at $s=0$ and the tenth at $s=9$.
At $s=24$, the views stay isolated throughout.}
\label{fig:routing}
\end{figure*}

\paragraph{Readout weighting and training correspondence.}
Equivalently, one may add a scalar $\gamma$ to all non-fusion-key
logits rather than modifying the fusion-key logit:
\begin{equation}
    o_f(\gamma)=
    \frac{e^\gamma\sum_{j\in\mathcal V}e^{s_j}v_j
          +e^{s_f}v_f}
         {e^\gamma\sum_{j\in\mathcal V}e^{s_j}+e^{s_f}},
    \qquad \gamma=-\log K.
    \label{eq:supp-readout-weighting}
\end{equation}
This exposes query multiplicity as an implicit control over the
relative attention to visual evidence and the accumulated fusion
state. Under symmetric initialization, $K$ controls this weighting
rather than the number of distinct learned reader states; this is
the interpretation of the sweep in Figure~\ref{fig:routing}.
In the symmetric training formulation, each query receives
the same gradient $g$. A single tied embedding instead receives
$\nabla_\theta\mathcal L_{\mathrm{tied}}=K g$; scaling this gradient
by $1/K$ restores the original per-query updates under matched
optimizer settings. The shared classification head is unchanged.
For $K=2$ and width 1024, the equivalent implementation reduces
trainable parameters from 4,098 to \textbf{3,074}, a \textbf{25.0\%}
reduction, without adding any trainable parameters to the attention projections.

\paragraph{Parameter counts.}
Table~\ref{tab:supp-parameters-count} compares the two-slot
formulation and its successive equivalent implementations at
logical $K=2$ and width 1024. All variants use a single fusion
classification head. Sharing frozen Transformer slots reduces
total stored parameters by \textbf{49.9\%}, from 606.31M to
303.95M, while retaining both visual token streams.
Merging the symmetric queries further reduces trainable
parameters from 4.098K to 3.074K. Both optimizations preserve
the model's predictions under the equivalence conditions above.

\begin{table}[!htbp]
\centering
\caption{Parameter counts for equivalent implementations of ReMoT.
M and K denote $10^6$ and $10^3$ parameters, respectively.}
\label{tab:supp-parameters-count}
\small
\setlength{\tabcolsep}{2.5pt}
\begin{tabular}{lcc}
\toprule
Implementation & Total (M) & Trainable (K) \\
\midrule
Separate slots & 606.31 & 4.098 \\
Shared weights & 303.95 & 4.098 \\
Shared + merged queries & 303.95 & 3.074 \\
\bottomrule
\end{tabular}
\end{table}

\section{Additional parameter ablations}
\label{sec:supp-parameters}

Across the settings in Figure~\ref{fig:routing}, ReMoT is least
sensitive to the number of fusion queries. Varying $K$ from 1 to 4
yields eight-group averages between 90.95\% and 91.21\%, a spread
of only 0.26 percentage points. Under query symmetry, $K$ controls
the attention weight of the fusion state
(Appendix~\ref{sec:supp-equivalence}). The sweep shows that readout
accuracy is stable across these weighting choices.
Interaction timing has a larger effect. With nine initial isolated
blocks, the average reaches 91.21\%, up 1.11 percentage points from
90.10\% with immediate interaction ($s=0$). This gain suggests that
an initial period of independent processing helps each view develop
its own forensic cues before exchanging evidence. All tested settings
with $6\leq s\leq 12$ exceed 90\%, and every tested setting with
$s\geq 15$ falls below 89\%. This pattern supports an intermediate
transition that leaves enough layers for cross-view exchange.
The patch-radius sweep is less sensitive. Accuracy exceeds 90\%
for $w\in\{3,5,7\}$, with a modest drop at the widest tested radius
and a lower score of 89.78\% at $w=1$. These results support using
a local neighborhood with enough spatial context for cross-view
patch exchange.

\section{Robustness evaluation details}
\label{sec:supp-robustness}

We evaluate ReMoT and 11 baseline detectors (NPR, UnivFD,
FatFormer, SAFE, C2P-CLIP, AIDE, DRCT, AlignedForensics,
Effort, DDA, and OmniAID) on the original GenImage validation
set, Mirage-Test, and Chameleon. JPEG
re-encoding at quality factors $\{90,80,70,60\}$ and Gaussian
blur with radii $\{0.5,1.0,1.5,2.0\}$ are applied independently
to decoded RGB images, before each detector's resizing, cropping,
and normalization. Both real and generated images undergo the
same processing, using Pillow's default JPEG encoder and
\texttt{GaussianBlur} filter. Blur radii are measured in source-image
pixels, so their relative strength depends on image resolution.
Together with the unmodified baseline, this gives nine conditions
per method and benchmark. We retain the fixed threshold of $0.5$
and the subset averaging protocol of Appendix~\ref{sec:supp-protocol}.
``None'' in the JPEG plots denotes the baseline condition,
not re-encoding at quality 100.

\end{document}